# Un modèle de base de connaissances terminologiques

P. Séguéla* - N. Aussenac-Gilles**

**Résumé - Abstract**

Dans cet article, nous défendons l'hypothèse que les Bases de Connaissances Terminologiques (BCT) sont d'autant plus utiles pour des utilisations différentes qu'elles ne sont pas formelles, et qu'elles se limitent au rôle de clarifier la terminologie d'un domaine en illustrant son utilisation en contexte. Nous avons mis au point un modèle de structuration de BCT afin qu'elles répondent à cet objectif. Ce modèle définit trois types d'entités reliées entre elles : termes, concepts et textes. Il rend également compte des particularités d'usage des termes. Les concepts sont représentés à l'aide de frames dont la description, non formelle, est normalisée. Associés à ce modèle, nous avons fixé des critères de modélisation des concepts. Nous montrons comment ces propositions contribuent au débat sur la situation des BCT par rapport aux ontologies et leur utilisation pour le développement d'applications en IA.

In the present paper, we argue that Terminological Knowledge Bases (TKB) are all the more useful for addressing various needs as they do not fulfill formal criteria. Moreover, they intend to clarify the terminology of a given domain by illustrating term uses in various contexts. Thus we designed a TKB structure including 3 linked features : terms, concepts and texts, that present the peculiar use of each term in the domain. Note that concepts are represented into frames whose non-formal description is standardized. Associated with this structure, we defined modeling criteria at the conceptual level. Finaly, we discuss the situation of TKB with regard to ontologies, and the use of TKB for the development of AI systems.

* CEA - CEN Cadarache - seguela@cea.fr
** IRIT - CNRS Toulouse - aussenac@irit.fr

**1. Introduction**

De définition récente, les Bases de Connaissances Terminologiques (BCT) répondent à un double besoin : mieux stocker des terminologies volumineuses puis faciliter leur utilisation pour différents types d'applications (Meyer 92) (Condamines 93). Elles viennent répondre à des problèmes de communication. En clarifiant le vocabulaire d'une communauté, elles permettent d'améliorer la pertinence des échanges entre agents et la gestion de l'information, tout en facilitant l'acquisition et la compréhension des connaissances ou l'exploitation de documents (Bourigault 95).

Les BCT constituent un enrichissement significatif des terminologies papier traditionnelles car elles comportent une trace des informations conceptuelles relevées par le terminologue au moment de l'identification des termes. Les BCT s'appuient sur un nouveau modèle de structuration des connaissances terminologiques qui différencie un niveau linguistique d'un niveau conceptuel. On accède ainsi par les termes du domaine à une modélisation conceptuelle. Cette composante s'apparente à la partie descriptive des bases de connaissances en IA, ou, encore plus étroitement, aux modèles du domaine et aux ontologies en ingénierie des connaissances (Aussenac 95). De ce fait, la notion de BCT, encore mouvante, fait l'objet des réflexions conjointes de chercheurs en IA et de linguistes terminologues[1] (Bourigault 95). Leur constitution soulève un ensemble de problèmes théoriques fondamentaux, au coeur des sciences cognitives (Bachimont 96). Il s'agit de définir un ensemble de primitives permettant d'associer avec pertinence les réseaux terminologiques et conceptuels, c'est-à-dire garantissant une bonne définition des concepts et une association valide avec les termes. Ce problème comporte deux facettes : comment définir mais aussi comment représenter - et valider- des concepts à partir de leur trace dans des textes. Nous avons choisi d'aborder ces questions par le biais des formalismes susceptibles de convenir pour construire une BCT (Séguéla 96) (Napoli 95).

---

[1] En France, le groupe Terminologie et Intelligence Artificielle du PRC-IA rassemble des chercheurs de ces deux communautés autour de la problématique des BCT (Bourigault 95).

Dans cet article, nous exposons un modèle de BCT mis au point en collaboration avec des terminologues[2] (Séguéla 96). A partir d'un état de l'art, nous étudions tout d'abord les notions à la base des BCT et précisons les données à représenter. Nous proposons ensuite un modèle de structuration de BCT définissant trois types d'entités reliées entre elles : termes, concepts et textes. Ce modèle rend aussi compte des particularités d'usage des termes. Associés à ce modèle, nous avons fixé des critères de modélisation des concepts. Pour conclure, nous montrons comment ces propositions contribuent au débat sur la situation des BCT par rapport aux ontologies et leur utilisation pour développer des applications en IA. En effet, nous défendons l'hypothèse que les BCT sont d'autant plus utiles pour des utilisations différentes qu'elles ne sont pas formelles, et que leur rôle se limite à clarifier la terminologie d'un domaine en illustrant son utilisation en contexte.

## 2. Quelques bases de connaissances terminologiques

Les BCT déjà développées sont peu nombreuses. Nous présentons ici trois projets parmi les plus avancés. Ils illustrent divers choix possibles de primitives de modélisation et de formalismes pour représenter les connaissances linguistiques et conceptuelles.

### 2.1 COGNITERM

Le projet COGNITERM (Meyer 92) a permis de construire une base de connaissances terminologiques bilingues dans le domaine des technologies de stockage optique[3]. Pour la première fois, un logiciel de gestion des connaissances était exploité pour représenter la terminologie d'un domaine et pour associer des données linguistiques à un réseau conceptuel (Skuce 91). Le système utilisé, CODE (Skuce 94), se veut adaptable. Il aide à organiser des connaissances en un réseau taxinomique de frames. CODE peut servir à définir, comprendre et présenter des connaissances (Lethbridge 94). Ce projet a nécessité une réflexion méthodologique sur l'utilisation de CODE par les terminologues. Récemment, cette équipe a développé un outil d'aide à l'extraction de terminologie à partir de textes qui peut servir pour identifier termes et concepts (Kavanagh 96).

Le modèle de données correspondant propose trois primitives : concepts, propriétés et termes. Les concepts sont définis par les relations taxinomiques qu'ils entretiennent avec d'autres concepts (ces relations formelles sont les primitives logiques du système et possèdent une sémantique associée) ainsi que par des relations non formalisées, les propriétés. Les données conceptuelles sont donc structurées et organisées en un réseau hiérarchique. L'héritage des propriétés permet de vérifier les associations conceptuelles et évite de dupliquer les définitions à tous les niveaux de concepts. A chaque concept, sont associés des termes, traités comme des entités indépendantes. Cependant, le texte n'est pas représenté dans le modèle de données de CODE. De plus, ce système général de gestion des connaissances propose au terminologue beaucoup plus d'informations qu'il n'en requiert pour construire une BCT. Par contre, étant basé sur des frames, CODE est bien adapté à la modélisation conceptuelle de la BCT. L'indépendance entre termes et concepts est intéressante en vue de fixer les descriptions des unités lexicales du domaine, de gérer polysémie et synonymie.

**2.2 Une BCT en géomorphologie glaciaire**

Afin d'évaluer l'aide que peut apporter l'étude de la terminologie d'un domaine à la modélisation des connaissances pour réaliser des SBC, une BCT a été réalisée dans le domaine de la géomorphologie glaciaire (Capponi 95). L'hypothèse à la base de ce travail était que la BCT serait d'autant plus facile à exploiter pour construire un SBC que les connaissances terminologiques et notamment conceptuelles y seraient formalisées. Dans ce projet, la construction de la BCT s'est appuyée sur les résultats du logiciel LEXTER (Bourigault 94) dont l'analyse du corpus a permis de relever les termes et concepts liés à la description du relief terrestre dans les glaciers.

---

[3] COGNITERM est le fruit de la collaboration de terminologues et d'informaticiens qui ont eu l'idée de rapprocher les résultats d'acquisition des connaissances et d'analyse terminologique, de manière complémentaire aux réflexions sur les ontologies en IA.

Le modèle de données différencie les concepts et les termes. Les classes de concepts sont décrites à l'aide d'un langage formel puis organisées en réseau sémantique. Elles sont caractérisées par leur relations vers d'autres concepts. Des informations linguistiques sont associées à chaque concept par l'intermédiaire de méta-instances (définition en langage naturel, termes associés, synonymes ou nombre d'occurrences dans le corpus). Ce réseau conceptuel et ces informations linguistiques forment la BCT.

Le formalisme retenu est une Logique Descriptive (ou Logique Terminologique), CLASSIC, afin d'assurer la cohérence logique de la couche conceptuelle de la BCT (Brachman 91). CLASSIC, issu de la famille KL-ONE, offre un langage restreint pour décrire les individus et classes d'individus d'un domaine et pour automatiquement les organiser en taxinomie d'après leur description formelle avec les primitives du langage. Le but de ce type de système est de proposer un langage terminologique minimal le plus expressif possible tout en garantissant la complétude et une complexité polynomiale de son algorithme de classification (Napoli 93).

## 2.3 Une BCT dans le domaine spatial

Nous avons également étudié la BCT bilingue modélisant la terminologie du domaine spatial présentée dans (Condamines 93). Construite dans le but de valider son intérêt dans le cadre d'un poste d'aide à la traduction et à la rédaction de documents, cette BCT se veut aussi un support à la présentation du domaine aux ingénieurs débutants ou encore un moyen de faciliter les échanges d'information dans l'entreprise[4]. En effet, cette terminologie a été envisagée comme un moyen de mettre en évidence les différences de vocabulaire et de points de vue entre métiers, et, ainsi, de parvenir à une représentation consensuelle faisant cohabiter ces points de vue et ces 'cultures'.

---

[4] Cette BCT a été réalisée dans le cadre du projet EUROLANG sur l'aide à la traduction. Elle est aujourd'hui intégrée à une plate-forme de Gestion Electronique de Documents à Matra Marconi Space.

L'originalité du modèle organisant les données terminologiques est de s'articuler autour de trois entités : à coté du concept et du terme, le lien concept/terme est isolé (Amsili 93). Au concept sont associées, entre autres, une définition en langage naturel, sa source et les relations qu'il entretient avec d'autres concepts. Les concepts sont organisés en un réseau structuré par trois relations (et leurs inverses) dont la sémantique n'est pas formelle : *est-générique-de*, *est-partie-de et est-élément-de*. Le terme comporte des informations linguistiques classiques : langue d'appartenance, catégorie grammaticale, nombre, genre, formes abrégées et formes elliptiques. Enfin, la relation concept/terme précise des informations sur le métier et l'usage, c'est-à-dire le contexte d'utilisation de ce terme pour désigner ce concept. Les données terminologiques, d'abord notées sur des fiches, sont ensuite complétées par les données conceptuelles avant d'être stockées dans une base de données. L'ensemble peut être consulté sous forme de graphes.

Cette BCT modélise les connaissances véhiculées par les langues de spécialité du domaine spatial et fixe ainsi la référence des termes. Cependant, elle ne contient pas de lien direct vers le corpus étudié. L'absence de ces liens est compensée en indiquant la source dans chaque concept ainsi qu'en faisant du lien concept-terme une entité à part entière, renvoyant aux différents usages du terme. Une autre caractéristique de cette approche est de se placer essentiellement du point de vue du terminologue, en ne se préoccupant ni de la cohérence ni de la consistance du réseau conceptuel, et en n'anticipant aucun type d'application pour laquelle la BCT serait construite a priori.

## 3. Les données d'une BCT

Une BCT est donc avant tout un inventaire des termes d'un domaine, enrichi d'informations conceptuelles permettant de donner un sens à ces termes, de définir les notions qu'ils désignent et de justifier leur place dans la terminologie. Sa structure sépare donc les donnés linguistiques, correspondant aux termes et textes, du réseau conceptuel, formé ses concepts et relations conceptuelles. Nous présentons ces deux composantes avant d'évoquer les problèmes soulevés par leur association.

### 3.1 Le réseau terminologique

### 3.1.1 Le corpus

Le domaine d'une terminologie est difficilement définissable dans l'absolu même si on arrive à le délimiter intuitivement par une description informelle liée à un produit technique, un ensemble de tâches, de métiers, etc. En pratique, le terminologue repère les termes du domaine dans un ensemble de textes préexistant à son analyse qu'il organise en corpus. Stocké sur support informatique, le corpus constitue à la fois une référence représentant les connaissances et le langage des spécialistes du domaine, et une source d'informations terminologiques (Bourigault 95). On appelle langue de spécialité la langue utilisée par une classe d'acteurs du domaine (i.e. un corps de métier) pour le décrire (Condamines 93). Cette langue comporte des termes (mots déviants de la langue générale, qui forment la terminologie) et des mots de la langue générale. On peut donc considérer que c'est ce corpus qui délimite le domaine et qu'il en offre, par extension, une description explicite. Cerner le domaine revient alors à constituer le corpus.

Si l'étude terminologique répond à un problème de communication entre experts, les ambiguïtés seront résolues en fixant les langues de spécialités du domaine, qui regroupe alors les activités individuelles ou collectives des experts. Ici, le corpus rassemble des textes les décrivant. Si la BCT est construite pour mieux définir le vocabulaire d'un domaine où les termes utilisés sont peu connus, le corpus sera plutôt formé de documents techniques. Le but de l'étude terminologique conditionne donc aussi le choix du corpus.

### 3.1.2 Les termes

Pour caractériser le langage des experts, le fonctionnement des termes utilisés est décrit en dehors du contexte du discours, en explicitant un modèle conceptuel du domaine auxquels ces termes réfèrent dans leur langue de spécialité. Les terminologues font l'hypothèse qu'au sein d'une langue de spécialité, un terme renvoie majoritairement à un concept unique (Bourigault 95). Les langues de spécialité permettent donc de réduire les situations de polysémie.

Ainsi, un terme est une manifestation linguistique d'une notion ou d'un concept repéré dans le discours des experts, dont il peut être considéré comme le label. Le terme est donc un signe linguistique qui se distingue du mot de la langue par sa fonction de référence à une notion du domaine. De ce fait, dans la plupart des modèles, aucune information conceptuelle n'est associée au terme. Ce terme peut être un syntagme verbal ou, le plus souvent, nominal. En effet, les études terminologiques visent en général à définir les désignations, déviantes et spécifiques, des objets d'un domaine. Elles ne s'attachent que plus rarement à dénominer les actions sur ces objets, qui sont plus consensuelles et moins spécifiques au domaine.

### 3.1.3 Les relations termes-textes

Dans une BCT, il est intéressant de conserver le corpus. Cet échantillon de textes constitue un élément de justification et d'explication du choix du terme et de ses liens vers des concepts. De plus, il peut alors être consulté par l'intermédiaire du réseau conceptuel.

Le lien terme-texte le plus naturel consiste à associer systématiquement le terme à ses occurrences dans le texte. Le terme devient alors une étiquette du lien concept-texte et la BCT représente une indexation "sémantique" de ce corpus. Cependant, l'interprétation de l'occurrence d'un syntagme dans un texte est un processus complexe, qu'un simple lien ne peut refléter. Dans le cas où un terme possède plusieurs sens (il est alors relié à plusieurs concepts), les différentes occurrences - ou contextes - doivent être regroupées en fonctions de ces sens. Afin de lever les ambiguïtés sur l'interprétation des termes, seules doivent être retenus les contextes caractérisant leur fonctionnement.

## 3.2 Le réseau conceptuel

### 3.2.1 Les concepts

Un concept est une structure permettant de décrire une notion à laquelle les experts ont fait référence via un ou plusieurs termes. Les concepts sont créés par le terminologue à partir de l'analyse des termes du corpus : leur sélection est donc guidée par le choix du corpus. Un concept est repéré par un identifiant qui, théoriquement, n'est pas un terme puisqu'il se peut que plusieurs termes servent à le désigner. Si on utilise un terme comme label, il doit avoir un statut particulier.

Le concept véhicule un ensemble d'informations sémantiques, sa description, qui doit faire apparaître des caractéristiques de la notion correspondante, telles que le terminologue les a repérées dans les textes. Cette description doit être suffisamment explicite, puisque la définition du terme se trouve dans l'interprétation de la description du concept associé (Bachimont 96). En fait, la description des concepts soulève trois types de problèmes : (a) le choix des critères de différenciation des concepts (quand va-t-on décider de créer un nouveau concept ?) (b) la gestion des différentes descriptions possibles d'un concept (ou points de vue) (c) l'interprétation de cette description, liée à son degré de formalisation.

### 3.2.2 La description des concepts

(a) Le modèle doit proposer des critères, linguistiques ou sémantiques, de création et de représentation des objets du domaine. Selon un critère linguistique, on créera un concept chaque fois que l'analyse linguistique du corpus fera émerger une notion absente du réseau conceptuel. Ainsi, si des termes différents sont utilisés pour désigner des notions proches, des concepts seront différenciés. Un autre critère linguistique est le patron syntactico-sémantique, qui caractérise les usages pour lesquels un terme désigne un certain concept. Différents patrons pour un même terme justifient la création de plusieurs concepts. Par contre, dans une perspective sémantique, on créera un concept chaque fois que l'interprétation du texte laisse supposer la définition ou la référence à un concept différent, sans qu'une justification linguistique confirme cette analyse.

(b) La description d'une notion du domaine est difficilement exhaustive et unique. De même qu'il existe différents mots pour désigner une notion, il existe plusieurs façons de la décrire. Pour les gérer, on peut associer au concept une classe d'équivalence regroupant plusieurs descriptions, nécessaires ou suffisantes, qui, ensemble, le décrivent de façon complète. Au contraire, une seule description peut unifier ses différentes caractéristiques et fonctionnalités selon les points de vue des experts.

(c) Plus on souhaite une description précise, d'interprétation non ambiguë, plus on a intérêt à en contraindre l'expression dans un langage formel. Il faut énoncer clairement si l'on prévoit que les concepts soient interprétés par un système informatique ou par des individus. Ainsi, on peut décrire les concepts par un texte en langage naturel, de façon normalisée ou formellement. Le degré de formalisation des descriptions conceptuelles est un problème lié aux objectifs de construction (donc aux utilisations prévues) de la BCT.

Si on choisit un langage formel, comme CLASSIC (Capponi 95), la définition d'un concept est liée à la sémantique formelle du langage (Sowa 91). La cohérence et la consistance de la couche conceptuelle de la BCT peuvent être vérifiées. De plus, la BCT peut fournir une partie de la couche domaine d'un SBC et en simplifier la construction. Par contre, le formalisme et surtout les connaissances décrites dans la BCT risquent de dépendre complètement du système à concevoir, ce qui va à l'encontre de la volonté d'une modélisation indépendante d'utilisations futures. Ainsi, des obstacles pratiques mais aussi théoriques s'opposent à l'exploitation de descriptions conceptuelles formelles dans des cadres applicatifs pour lesquels elles n'ont pas été prévues (Charlet 94). Enfin, ce caractère formel limite la souplesse de la structuration de la BCT, et réserve sa construction et sa consultation aux utilisateurs familiers des langages formels.

A priori, à la différence des concepts présents dans une ontologie, il n'est donc pas forcément utile que les concepts d'une BCT soient formels. Cependant, si on souhaite exploiter une BCT pour construire une ontologie du domaine ou un SBC, on a tout intérêt à anticiper les exigences en formalisation et à prévoir des descriptions normalisées. Normaliser un concept, c'est fixer une référence de définition et d'interprétation, par exemple en structurant sa description en langage naturel ou au moyen d'une structure comportant des propriétés. Ces descriptions restent facilement interprétables par leurs utilisateurs. De plus, en les organisant dans une structure taxinomique comme celle de CODE, on peut garantir leur cohérence par un processus d'héritage (Skuce 91).

### 3.2.3 Les relations conceptuelles

Les relations conceptuelles ont comme double objectif de décrire les concepts et de déclarer les observations sur ces concepts révélés par l'analyse des textes. On distingue les relations structurelles ("est-un", "partie-de") des relations assertionnelles (traduisant des faits). Elles favorisent une description statique des objets du domaine, puisque l'objectif de modélisation d'une BCT n'est pas lié au raisonnement. Le nombre et la nature des types de relations qui peuvent être utilisés dans une BCT influe sur les possibilités de formalisation. Leur type peut être défini soit en fonction du domaine soit a priori. Dans ce cas, la sémantique doit être assez précise pour qu'on puisse composer les relations et interpréter les descriptions conceptuelles en les parcourant.

Sa sémantique étant bien définie, on choisit généralement comme relation structurelle la relation "est-un". Elle organise les concepts en une taxinomie, qui est exploitée grâce à un système d'inférence basé sur l'héritage des propriétés. Pour définir d'autres relations structurelles, il est souvent plus facile de tenir compte des particularités du domaine que de formaliser a priori une relation générique. De plus, il est risqué de combiner l'usage de plusieurs relations structurelles formelles, car alors, la structure taxinomique du réseau conceptuel disparaît et, avec elle, ses propriétés.

D'autres relations, non structurelles, sont possibles. Mais le but de la BCT n'est pas d'expliciter tous les faits entre les concepts du domaine. Ne seront considérées comme pertinentes pour la modélisation que les relations traduites par des relations différentielles entre termes. Ces relations assertionnelles complètent la description du domaine et précisent la définition des concepts, en plus des relations structurelles et des propriétés.

## 3.3 Du réseau linguistique au réseau ontologique

### 3.3.1 Le lien terme-concept

Dans la BCT, le terme est un moyen d'accès au concept mentionné par les experts du domaine et relevé par le terminologue lors de son étude. Cependant, la désignation du concept n'est pas systématiquement consensuelle dans le langage des experts. Ainsi, selon son métier et le contexte du discours, un locuteur peut se référer à un concept ou un autre avec le même terme (Condamines 93).

En effet, dans une langue de spécialité, il est rare qu'un terme désigne exclusivement un concept (Rebeyrolle 95) : un ou plusieurs termes sont utilisés de façon préférentielle pour désigner un concept. La notion de point de vue vise à caractériser cette préférence et permet, par exemple, de parler d'équivalence ou de synonymie lorsque deux termes désignent, pour un même point de vue, le même concept. Il est alors envisageable que plusieurs concepts soient désignés par le même terme relativement à différentes langues de spécialité. Les différents points de vue sur le concept au sein d'une BCT sont repérés par des critères linguistiques, en observant les différentes langues de spécialité utilisées par les acteurs du domaine. Mais on considère qu'il n'est pas possible qu'un terme soit utilisé de façon préférentielle dans une langue de spécialité pour désigner deux concepts distincts. Cette affirmation, intuitivement correcte, est importante pour interpréter la BCT: la notion de point de vue permet ainsi d'interpréter les termes hors contexte sans équivoque. Un terme désignera un et un seul concept relativement à un point de vue.

### 3.3.2 Le lien terminologique - ontologique

C'est dans le lien entre la composante linguistique et le réseau conceptuel que se situe toute la problématique de la représentation des connaissances terminologiques. D'une part, les termes influencent fortement le réseau des concepts : la différenciation des concepts et la granularité de la modélisation sont guidés par les termes et donc par le corpus. D'autre part, la finalité de la partie conceptuelle de la BCT vient contraindre aussi cette modélisation : suivant le degré de formalisation des concepts, la sémantique du réseau conceptuel reflète celle des textes ou s'adapte au raisonnement formel. Dans tous les cas, il faut spécifier la sémantique des primitives (concepts et relations). Ainsi, on doit pouvoir vérifier qu'un concept est défini de manière unique, cohérente par rapport aux autres concepts, et que sa signification n'est pas modifiée suivant le contexte, lorsqu'on le combine à d'autres primitives.

Bachimont met en avant la sémantique différentielle de la langue pour poser comme préalable à la construction d'une ontologie formelle la représentation d'un réseau de termes où sont explicitées leurs différences sémantiques (Bachimont 96). Dans ce but, il faut garantir que chaque définition de terme soit unique, cohérente et non contextuelle. Pour cela, on doit fixer un point de vue lié à une classe d'applications, qui assure une sémantique différentielle au réseau conceptuel. Dans le cadre d'une BCT, l'objectif est moins contraignant : il suffit que la normalisation des définitions de concepts rende compte des différences entre les notions exprimées par les termes, sans prétendre s'adapter à une application.

**4. Proposition d'un modèle de BCT**

Nous considérons que l'intérêt d'une BCT est de fixer la signification des termes du domaine. Nous présentons un modèle qui permet une structuration normalisée des données terminologiques pour en assurer une interprétation non ambiguë et précise, et pour faciliter plusieurs types d'utilisation de la BCT.

**4.1 Le modèle des données**

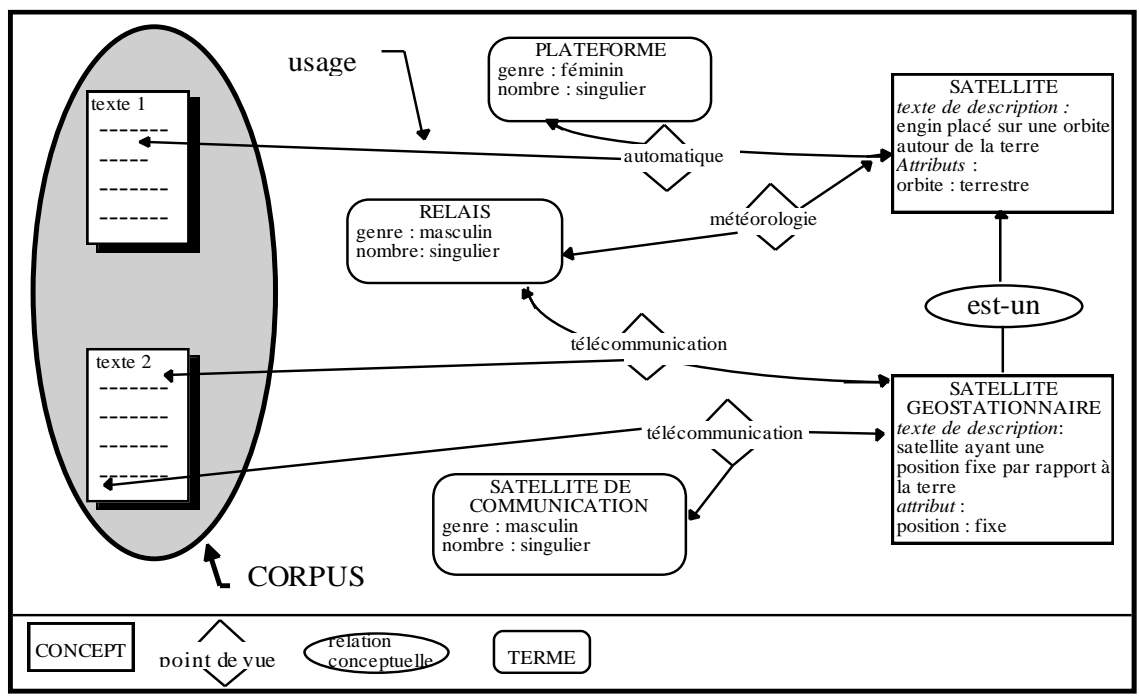

**Figure 1 : Les différents composants d'une BCT selon le modèle proposé**

Nous proposons un modèle de BCT formé de 3 types de structures : textes, termes et concepts (Séguéla 96), dont les liens véhiculent également des informations :

- les liens d'usage précisent chacune des occurrences d'un terme d'une langue de spécialité en reliant le lien (terme, concept) à des textes ;
- les points de vue spécifient la validité d'utilisation de la désignation conceptuelle du terme ; il est une des caractéristiques du lien terme-concept ;
- des relations sémantiques associent les concepts entre eux.

La figure 1 illustre l'organisation de ces structures. Dans cet exemple, pris dans le domaine spatial (Condamines 93), le terme RELAIS possède deux interprétations. Dans la langue de spécialité des experts en météorologie (point de vue *météorologie*), il désigne le concept étiqueté par SATELLITE et décrit comme " engin placé sur une orbite autour de la terre ". Le terme RELAIS est aussi utilisé dans la communauté des télécommunications pour désigner le concept étiqueté SATELLITE GEOSTATIONNAIRE. D'ailleurs, selon ce point de vue, les termes RELAIS et SATELLITE DE COMMUNICATION sont synonymes.

### 4.1.1 Textes et termes

Les textes sont stockés dans la BCT sous forme d'unités textuelles de manière à faciliter leur gestion. Ce découpage est transparent pour les utilisateurs.

A priori, les termes présents dans la BCT doivent être utilisés au moins une fois dans le corpus, exceptés ceux mentionnés au cours d'entretiens par exemple. Cependant, les termes ne sont pas directement reliés aux textes, car plusieurs interprétations possibles du terme sont possibles au sein de la BCT. Afin d'éviter ces ambigüités, les textes sont reliés au lien terme-concept, illustrant un usage du terme pour lequel il désigne ce concept. Les termes peuvent être repérés dans les textes manuellement ou bien à l'aide d'un logiciel d'extraction de terminologie comme LEXTER (Bourigault 94).

Afin de marquer la différence entre le signe linguistique et les notions qu'il désigne, on n'associe au terme aucune information conceptuelle. La signification d'un terme lui est donnée par les concepts associés et son interprétation dépend de la sémantique des structures de représentation des concepts. Il est donc important de définir conjointement termes et concepts de la BCT. Pour déterminer si deux termes sont proches, assimilables ou différents, on fait appel au concept associé. Les relations entre termes, comme la synonymie, la polysémie ou l'anaphore, sont implicites et calculables à partir des liens terme-concept. Seules les relations grammaticales (morphologiques ou de composition i.e. 'est-en-tête-de' (Bourigault 94)) doivent être explicitées. La structure de terme, proche de celle présentée dans (Amsili 93), rassemble donc, en plus du syntagme qui le désigne, uniquement des informations linguistiques : langue, variantes de forme, décomposition grammaticale, genre, nombre, ...

**4.1.2 Concepts et relations conceptuelles**

Afin d'organiser les concepts en réseau de manière cohérente et surtout d'assurer que leur interprétation soit conforme à celle des concepteurs de la BCT, nous avons opté pour une description structurée et normalisée des concepts. Les concepts sont représentés par des frames qui permettent d'expliciter les connaissances de différenciation et des relations assertionnelles et structurelles vers d'autres concepts. Les informations décrites dans ces structures suivent donc un format précis, mais non interprétable par la machine :

- un identifiant choisi parmi les termes associés, qui sera alors un "terme-vedette";
- une description, texte qui permet d'associer librement des informations au concept ;
- un ensemble d'attribut:valeur, attributs et valeurs étant des chaînes de caractères ;
- des relations assertionnelles vers d'autres concepts, sans sémantique formelle, définies a priori ou en fonction du domaine ;
- des relations structurelles " est-un " vers des concepts subsumants : un concept hérite des attributs et des relations assertionnelles des concepts subsumants.

Un concept est décrit par sa place dans la hiérarchie des concepts. Sa définition est liée à l'interprétation de ses attributs, de leurs valeurs et de ses relations structurelles. Elle est enrichie par les relations non structurelles, qui peuvent être interprétées sans erreur grâce à un texte de définition. Comme la modélisation conceptuelle de la BCT vise la caractérisation des notions et non leur définition exhaustive, un concept, selon un point de vue particulier, est caractérisé de manière unique par ses propriétés et ses relations avec d'autres concepts. Si besoin, il faut donc définir autant de concepts que de descriptions différentes de celui-ci.

### 4.1.3 Critère de modélisation

Le lien terme-concept comporte des informations qui caractérisent son champ de validité, c'est-à-dire dans quelles conditions ce terme désigne ce concept :

- des liens vers des textes illustrant des usages du terme;
- des points de vue, indiquant des groupes de locuteurs (des langues de spécialité).

Nous retenons des critères de modélisation essentiellement linguistiques. Ainsi, un nouveau concept est créé chaque fois que l'analyse des textes marque la différentiation d'un nouveau concept. Par exemple, l'utilisation d'un terme, reflétée par un de ses patrons syntactico-sémantiques, traduit un point de vue différent sur un concept existant. Ce sont les relations conceptuelles, les attributs et leurs valeurs qui vont représenter les différences entre concepts. D'après ce choix, les relations structurelles ont, en plus du rôle d'organisation du modèle, un rôle de différenciation linguistique des concepts reliés.

### 4.2 Un environnement de saisie et de consultation de la BCT

A partir de ce modèle, un environnement de gestion de la BCT a été spécifié de manière à disposer des avantages offerts par la structuration des données, tout en laissant ce modèle transparent pour les utilisateurs. Cet environnement est en cours de conception. Afin de privilégier l'accès à des terminologies non formelles mais normalisées, exploitées par des individus et non par un système formel, nous avons opté pour un système de gestion de base données, auquel sont associés des éditeurs, des fonctionnalités de consultation et de vérification ainsi qu'un éditeur de graphes. Le tout sera disponible sur PC et station de travail en JAVA.

**4.2.1 Construction de la BCT**

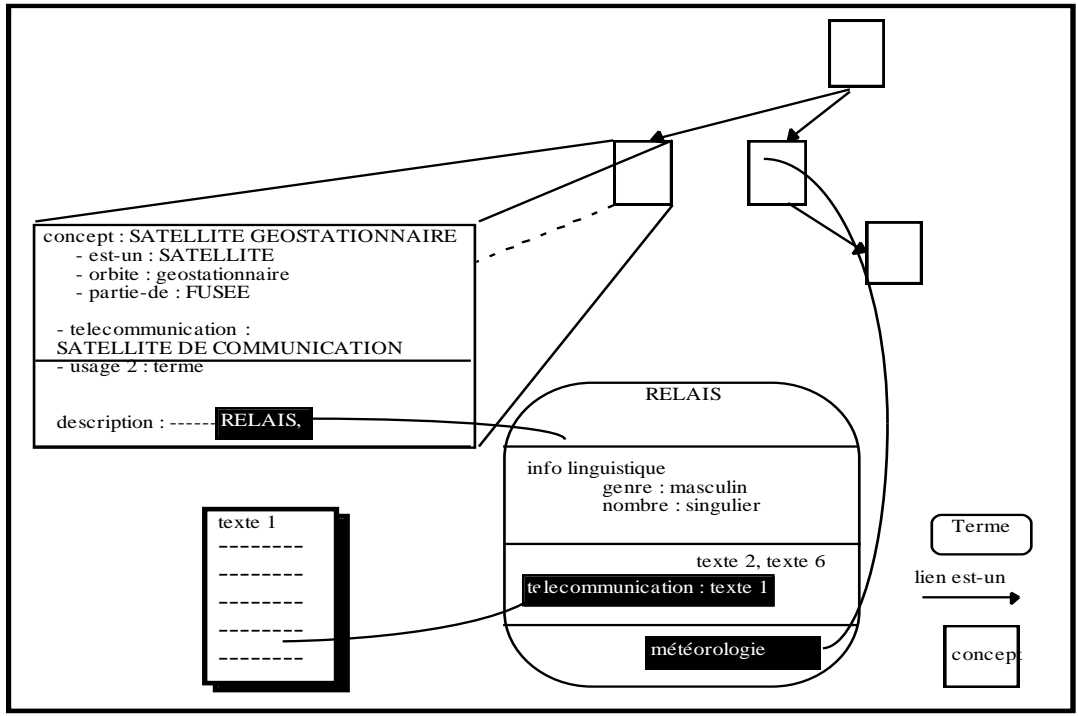

**Figure 2 : La consultation d'une BCT via l'interface graphique**

L'environnement de saisie de la BCT doit être convivial et simple. Destiné au terminologue qui dialogue avec l'expert, il doit prendre en compte sa démarche. Le corpus doit pouvoir être augmenté de nouveaux textes formatés, qui seront reliés aux liens termes-concepts. Les termes doivent pouvoir être inventoriés et décrits au fur et à mesure de l'analyse des textes. Cette analyse implique aussi la description, sous une forme normalisée, de concepts, qui sont placés dans la hiérarchie et reliés aux concepts déjà définis pour déclarer des connaissances. Enfin, plusieurs fonctionnalités, comme l'éditeur graphique, permettent au terminologue de vérifier son travail.

**4.2.2 Consultation de la BCT**

Ainsi définie, cette BCT convient pour un utilisateur qui souhaite accéder à un lexique informatisé, suffisamment riche pour lui indiquer le sens des mots du domaine, pour lever les problèmes de polysémie et de synonymie, et lui fournissant des exemples d'utilisation des termes dans des textes. Par contre, elle ne prétend pas fournir des descriptions ontologiques directement utilisables dans des systèmes formels. Les accès prévus à son contenu sont inspirés du système décrit dans (Amsili 93) :

- Un accès graphique (figure 2) par le réseau conceptuel : Suivant les types de relations affichées, on visualise la hiérarchie ou le réseau des relations assertionnelles entre concepts. A partir des nœuds-concepts, désignés par leur identifiant, on peut consulter leur description normalisée, connaître les termes les désignant pour différents locuteurs et accéder aux informations associées aux termes.
- Un accès depuis le corpus des textes, dans lequel les termes identifiés sont mis en valeur ; on envisage également un accès direct au texte par mot clé.
- Un accès par les listes de termes et de concepts présentés par ordre alphabétique.

## 5. Enjeux liés à l'utilisation d'une BCT

Pour conclure, nous souhaitons revenir sur les questions soulevées par l'utilisation des connaissances que renferme une BCT pour construire des applications, en particulier pour décrire les objets manipulés par un système d'IA. Nous soulignons la différence importante d'objectifs entre les représentations d'un domaine qu'offrent les BCT telles que nous les définissons, les ontologies et les modèles du domaine.

### 5.1 BCT et modèles du domaine

En ingénierie des connaissances, dans un modèle du domaine, il s'agit de représenter les objets nécessaires pour spécifier la tâche assignée à un système d'IA. Leur définition doit être suffisamment précise pour donner naissance à des objets formels exploitables sans ambiguïté par le système opérationnel. Ces objets peuvent être repérés à partir des connaissances d'experts ou de textes. Au contraire, la modélisation conceptuelle dans une BCT a pour but de structurer et différencier les objets ou référents désignés par les termes dans un domaine délimité par un corpus de textes. Les concepts émergent de l'analyse linguistique du corpus et de l'étude des termes qui servent d'identifiants conceptuels.

Dans le but de construire des résolveurs de problème, notre proposition rejoint l'analyse de B. Leroux, qui considère que l'unification terminologique permet de lutter contre l'hétérogénéité des modèles du domaine (Leroux 96). Pour lui, une BCT est une base pour constituer une ontologie d'un domaine en fonction d'une tâche particulière, et ensuite une BC en assignant à ces objets du domaine des rôles dans une méthode de résolution de problème adaptée à la réalisation de cette tâche. Les connaissances terminologiques doivent donc laisser la place à l'interprétation de leurs utilisateurs en pouvant s'adapter au type de problème d'une application et à sa formalisation.

**5.2 BCT et ontologies**

Une ontologie est définie comme une spécification explicite, au niveau des connaissances, d'une conceptualisation, ou encore comme l'ensemble des distinctions pertinentes pour un agent pour définir les concepts d'un domaine (Gruber 92) (Van Heijst 97). Ainsi, l'ontologie d'un domaine rassemble des objets décrits selon les besoins de la réalisation d'une tâche dans ce domaine (Leroux 96). Enfin, les ontologies génériques sont des modélisations conceptuelles réalisées indépendamment de leur utilisation et prétendant, par leur formalisation, opérationnalisation et sémantique associée, être réutilisables par différentes applications dans ce domaine (Guarino 95).

La définition d'une ontologie rencontre comme première difficulté le choix de critères descriptifs requis par le système formel. Ces critères doivent être valides a priori, indépendamment de tout point de vue sur le domaine. La description des concepts proprement dite pose un deuxième ensemble de problèmes : aussi exhaustive soit elle, on n'est pas sûr qu'elle suffise pour le bon fonctionnement d'un système opérationnel. Enfin, le choix des noms des concepts est délicat : du moment que les identifiants sont des mots du domaine, leur interprétation risque de refléter un point de vue.

Ainsi, la réutilisation de l'ontologie médicale UMLS pour la modélisation de la couche domaine d'un SBC se heurte à l'ambiguïté de l'interprétation sémantique de ses relations et à une mauvaise adaptation aux caractéristiques du système (ici MENELAS) prévu en aval (Charlet 94). Cette étude souligne que la structuration des connaissances du domaine selon une sémantique formelle, indépendamment de toute application, est indispensable pour leur bonne interprétation. Mais elle ne garantit par pour autant que l'ontologie puisse être utilisée par divers systèmes d'inférence. Ainsi, B. Bachimont remet en cause l'objectif de généricité dans la modélisation d'ontologies. Pour lui, seules sont réalisables des ontologies régionales du domaine, qui sont des modèles instanciés d'ontologies dans lesquels un point de vue est fixé pour assurer une propriété référentielle aux concepts.

**5.3 La prise en compte des points de vue dans la BCT**

L'indépendance de point de vue sur le domaine doit rester un objectif prioritaire pour définir des primitives de modélisation conceptuelle d'une BCT. Dans le modèle que nous proposons, la dissociation linguistique-conceptuel permet de rendre compte des points de vue. Les différentes façons de considérer ou d'utiliser un concept sont accessibles au niveau linguistique grâce à l'indication de locuteurs dans les liens terme-concept. Ainsi, l'utilisateur de la BCT est averti de l'existence de points de vue et peut y accéder. Par contre, la description du concept dépend de ses attributs mais aussi de ses relations avec les autres concepts, qui peuvent correspondre à plusieurs points de vue, sans que l'on sache ceux pour lesquels elles sont valides. Notre modèle permet donc seulement de préciser le point de vue selon lequel un concept est évoqué par un terme, sans différencier ses descriptions par point de vue. Une perspective actuelle de notre travail est d'affiner l'organisation du réseau conceptuel de manière à pouvoir associer, si nécessaire, un point de vue à une relation conceptuelle. Une piste possible serait d'introduire le référé comme objet visible selon différents points de vue via autant de concepts.

**5.4 Conclusion**

La mise au point d'un modèle de BCT nous a donc conduit à donner priorité au rôle de référentiel clarifiant la terminologie d'un domaine et rendant compte des points de vue et usages des termes dans le domaine. Nous distinguons bien notre perspective de celle des ontologies, qui, elles, privilégient une description formelle et réutilisable des notions d'un domaine. Pour cela, nous avons retenu une représentation normalisée mais non formelle des connaissances conceptuelles, les concepts étant des frames organisés en hiérarchie, décrits par des attributs et des relations assertionnelles. Nous avons également mis en valeur les points de vue en isolant les liens entre termes et concepts, et enrichi la description conceptuelle par un accès aux textes depuis les liens termes-concepts. Il nous reste à terminer le développement d'un environnement de saisie d'une BCT structurée selon ce modèle, de manière à valider ces choix. Dans ce but, nous participons à un projet en collaboration avec l'ERSS et EDF-DER. La BCT développée doit servir à construire un système de consultation de documents adapté à plusieurs classes d'utilisateurs dans le domaine du logiciel scientifique. Ce travail permettra de juger de la pertinence du modèle de BCT pour la construire et en exploiter le contenu pour concevoir une application.

## Références

Amsili, P., Condamines, A., Monier, P., Soubeille, A.M. (1993). *Modèle Terminologique pour Matra Marconi Space*, Rapport MMS/ARAMIIHS.

Aussenac-Gilles N., Bourigault D., Condamines A., Gros C. (1995) " How Can Knowledge Acquisition Benefit from Terminology ? ", *Proc. 9th Knowledge Acquisition for Knowledge-Based Systems Workshop*, Banff. 1/1-1/19.

Bachimont B. (1996). " Engagement sémantique et engagement ontoloqique : propositions méthodologiques et problèmes théoriques à propos des ontologies en IA ". *Actes des journées Acquisition des Connaissances*. Sète.

Bourigault, D., (1994). *LEXTER, Un Logiciel d'EXtraction de TERminologie. Application à l'Acquisition de Connaissances à partir de Textes*. Thèse de doctorat. EHESS, Paris.

Bourigault, D., Condamines, A., (1995). " Réflexions sur le concept de base de connaissances terminologiques ". Actes des Journées du PRC-GDR-IA, Nancy. TEKNEA, Toulouse. 425-444.

Brachman, R., McGuiness, D.L., Patel-Schneider, P.F., Borgida, L., (1991) " Living with Classic : When and How to Use a KL-ONE-Like Language ". *Principles of Semantic Networks*, J. Sowa ed., Morgan Kaufman, 401-457.

Capponi. N. (1995). *Modélisation d'une Base de Connaissance Terminologique*. Mémoire de DEA Informatique. Nancy. 1995.

Charlet, J., Bachimont, B., Bouaud, J., Zweigenbaum,P., (1994). " Ontologie et Réutilisabilité : Expérience et Discussion ". *5èmes Journées Acquisitions des Connaissances.* Strasbourg. C1-C14.

Condamines, A., Amsili P., (1993). " Terminologie, entre Langage et Connaissances : un Exemple de Base de Connaissances Terminologiques. *Terminology and Knowledge Engineering*, Frankfurt, 316-323.

Condamines, A., (1996). " Quelles bases Terminologiques pour les Nouveaux Besoins en Entreprise ? " *Terminogramme* n°78.

Gruber, T.R., (1992). " A Translation Approach to Portable Ontology Specifications ". In *Proc. of the Second Japanese Knowledge Acquisition Workshop*. Kobe and Tokyo.

Guarino, N., Giaretta, P. (1995). " Ontologies and Knowledge Bases : Towards a terminological Clarification ".*Towards very large Knowledge Bases : Knowledge Building and Knowledge Sharing.* Mars, N. (ed.). Amsterdam : IOS Press. 25-32.

Kavanagh, J., (1996). *The Text Analyzer : a Tool for Extracting Knowledge from Text*. Master's of computer science Thesis, Univ. of Ottawa (Can).

Leroux, B., (1996). " Eléments d'une classification des approches ontologiques ", *Actes des 6èmes Journées d'Acquisition des Connaissances*, Sète. 109-122.

Lethbridge, T., Skuce, D. (1994). " Knowledge Base Metrics and Informality: User Studies with CODE4 ". *Proc. 8th Knowledge Acquisition for KBS Workshop*, Banff. 10/1-10/19.

Meyer, I., Skuce, D., Bowke, L., Eck, K., (1992). " Toward a New Generation of Terminological Ressources : An Experiment in Building a Terminological Knowledge Base ". *Proc. 13th Int. Conference on Computational Linguistics*. 956-960. Nantes.

Napoli, A., Volle, P., (1993). *Une introduction aux logiques terminologiques*. Rapport de recherche 93-R-033, Centre de Recherche en Informatique de Nancy, Vandoeuvre.

Rebeyrolle, J., (1995). *Apports de la terminologie à l'étude des points de vue*. Mémoire de DEA en linguistique, ERSS, Univ. Toulouse Le Mirail.

Séguéla, P., (1996). *Formalisation et modélisation des connaissances terminologiques*. Mémoire de DEA en Informatique, IRIT, Univ. P. Sabatier, Toulouse. 55 p.

Skuce, D., Meyer, I., (1991). " Terminology and Knowledge Acquisition: Exploring a Symbiotic Relationship ". *Proc. 6$^{th}$ Knowledge Acquisition for KBS Workshop*, Banff. 29/1-29/21.

Skuce, D., Lethbridge T.C., (1994). " CODE4 : A Multifunctional Knowledge Management System ". Proc. *8$^{th}$ Knowledge Acquisition for Knowledge-Based Systems Workshop*, Banff. 12/1-12/21.

Sowa, J.F. (1991). *Principles of Semantic Networks*. Morgan Kaufman.

Van Heijst, G., Schreiber, A.Th., Wielinga, B.J., (1997). " Using Explicit Ontologies in KBS Development ". *International Journal of Human-Computer Studies*. , London : Academic Press. in press.